\pdfoutput=1

\documentclass[11pt]{article}

\usepackage
{acl}

\usepackage{times}
\usepackage{latexsym}
\usepackage{makecell}
\usepackage[T1]{fontenc}

\usepackage[utf8]{inputenc}

\usepackage{microtype}

\usepackage{hyperref}
\usepackage{booktabs}
\usepackage{graphicx}
\graphicspath{{./imgs/}}
\usepackage{footmisc}
\usepackage{microtype}

\usepackage{caption}
\usepackage{subcaption}
\usepackage{array, makecell} %

\usepackage{pifont}
\newcommand{\cmark}{\ding{51}}%
\newcommand{\xmark}{\ding{55}}%

\usepackage{tabularx,colortbl}

\usepackage{tikz}
\usepackage{collcell}
\usepackage{fdsymbol}
\usepackage{etoolbox}

\newtoggle{inTableHeader}
\toggletrue{inTableHeader}
\newcommand*{\StartTableHeader}{\global\toggletrue{inTableHeader}}%
\let\OldTabular\tabular%
\let\OldEndTabular\endtabular%
\renewenvironment{tabular}{\StartTableHeader\OldTabular}{\OldEndTabular\StartTableHeader}%

\newcommand*{\MinNumber}{-1.0}%
\newcommand*{\MidNumber}{0.0} %
\newcommand*{\MaxNumber}{1.0}%
\newcommand{\cmmnt}[1]{}

\newcommand{\ApplyGradient}[1]{%
  \iftoggle{inTableHeader}{#1}{
    \ifdim #1 pt > \MidNumber pt
        \pgfmathsetmacro{\PercentColor}{max(min(100.0*(#1 - \MidNumber)/(\MaxNumber-\MidNumber),100.0),0.00)} %
        \hspace{-0.33em}\colorbox{yellow!\PercentColor!blue}{#1}
    \else
        \pgfmathsetmacro{\PercentColor}{max(min(100.0*(\MidNumber - #1)/(\MidNumber-\MinNumber),100.0),0.00)} %
        \hspace{-0.33em}\colorbox{blue!\PercentColor!blue}{#1}
    \fi
  }}
\newcolumntype{R}{>{\collectcell\ApplyGradient}c<{\endcollectcell}}

\usepackage{amsmath}
\usepackage{amsfonts,bm}
\usepackage{xspace}

\newcommand{\blue}[1]{\textcolor{blue}{#1}}

\newcommand{\ie}{{\em i.e.,}\xspace}
\newcommand{\eg}{{\em e.g.,}\xspace}

\newcommand{\Ni}{({\em i})~}
\newcommand{\Nii}{({\em ii})~}
\newcommand{\Niii}{({\em iii})~}
\newcommand{\Niv}{({\em iv})~}

\definecolor{mypink3}{cmyk}{0, 0.7808, 0.4429, 0.1412}

\makeatletter   
\newcommand{\sveryshortarrow}[1][3pt]{\mathrel{%
    \vcenter{\hbox{\rule[-.5\fontdimen8\scriptfont3]
               {\scriptratio\dimexpr#1\relax}{\fontdimen8\scriptfont3}}}%
   \mkern-4mu\hbox{\let\f@size\sf@size\usefont{U}{lasy}{m}{n}\symbol{41}}}}
\makeatother

\def\eqref#1{equation~\ref{#1}}

\def\1{\bm{1}}

\def\vh{{\bm{h}}}

\def\vp{{\bm{p}}}

\def\vz{{\bm{z}}}

\def\m1{{\bm{1}}}

\def\mH{{\bm{H}}}

\def\mZ{{\bm{Z}}}

\DeclareMathAlphabet{\mathsfit}{\encodingdefault}{\sfdefault}{m}{sl}
\SetMathAlphabet{\mathsfit}{bold}{\encodingdefault}{\sfdefault}{bx}{n}

\def\gD{{\mathcal{D}}}

\usepackage[nameinlink]{cleveref}
\crefformat{section}{\S#2#1#3} 
\crefname{algorithm}{Alg.}{Algs.}
\crefformat{subsection}{\S#2#1#3}
\Crefname{equation}{Eq.}{Eqs.}
\Crefname{figure}{Fig.}{Figs.}

\usepackage[colorinlistoftodos,prependcaption,textsize=tiny]{todonotes}

\newcommand{\change}[1]{{\leavevmode\color{black}#1}}

\usepackage{soul}

\usepackage{float}

\usepackage{multirow}
\usepackage{hhline}

 \title{ChartQA: A Benchmark for Question Answering about Charts \\with Visual and Logical Reasoning}

\author{
Ahmed Masry$^{\clubsuit}$, \  Do Xuan Long$^{\spadesuit}$, \ Jia Qing Tan$^{\spadesuit}$,  \ Shafiq Joty$^{\spadesuit\vardiamondsuit}$, \ Enamul Hoque$^{\clubsuit}$\\
$^\clubsuit$York University, Canada \\
$^\spadesuit$Nanyang Technological University, Singapore, $^\vardiamondsuit$Salesforce Research 
\\
$^\clubsuit$\{masry20, enamulh\}@yorku.ca\\
$^\spadesuit$\{xuanlong001@e.ntu, C190022@e.ntu, srjoty@ntu\}.edu.sg 
}

\begin{document}
\maketitle

\begin{abstract} Charts are very popular for analyzing data. When exploring charts, people often ask a variety of complex reasoning questions that involve several logical and arithmetic operations. They also commonly refer to visual features of a chart in their questions. However, most existing datasets do not focus on such 
complex reasoning questions as their questions are template-based and answers come from a fixed-vocabulary. In this work, we present a large-scale benchmark covering 9.6K human-written questions as well as 23.1K questions generated from human-written chart summaries. To address the unique challenges in our benchmark involving visual and logical reasoning over charts, we present two transformer-based models that combine visual features and the data table of the chart in a unified way to answer questions. While our models achieve the state-of-the-art results on the previous datasets as well as on our benchmark, the evaluation also reveals several challenges in answering complex reasoning questions.

\end{abstract}

\section{Introduction}

\begin{figure}[t] 
\linespread{0.5}\selectfont\centering 
\renewcommand{\arraystretch}{1.0}
\scalebox{0.94}{\begin{tabular}{p{3.5cm}p{4.5cm}}     
 \raisebox{-0.5\height}{\includegraphics[width=3.5cm]{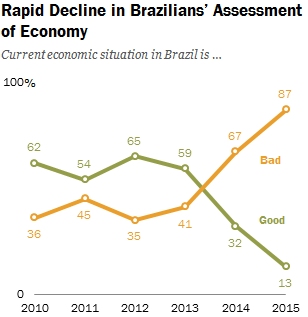}}
        &
         {\begin{tabular}{@{}p{3.9cm}@{}}
         {\small{\textbf{Q1:} \small Which year has the most divergent opinions about Brazil's economy?}}
         \\ [+2em]
         {\small{\blue{\textbf{Answer:} 2015}}}
         \\\\
         \small{\textbf{Q2:} \small What is the peak value of the orange line?} 
         \\ [+1.2em]
         {\small{\blue{\textbf{Answer:} 87}}}
         \end{tabular}}
    \end{tabular}
    }
    \caption{Sample 
    questions in our benchmark.
    }
    \label{tab:example}
\end{figure}

Data visualizations such as bar charts and line charts have become popular in analyzing data and making informed decisions. To analyze data, often people ask complex reasoning questions about charts involving arithmetic and logical operations~\cite{kim2020answering}. Answering such questions requires   
a significant amount of perceptual and cognitive efforts  as people need to combine multiple operations such as retrieving values, comparing values, finding maximum, calculating sums and differences 
of values. For example, the question Q1 in \Cref{tab:example} requires the user to compute the differences between the two lines for each year and find the year with the highest difference.

The goal of a Chart Question Answering (ChartQA) system is to help users by taking a chart and a natural language question as input and predicting the answer. 
This task differs from other  QA tasks such as QA on texts~\cite{squadv1} and tables~\cite{sempre} because the input for ChartQA is a visual representation of data 
that can draw a reader’s attention to various prominent features such as trends and outliers ~\cite{kim2020answering,kim2021towards}. Also, people tend to ask questions by referring to visual attributes of marks. For example, in \Cref{tab:example}, Q2 
refers to the color of a mark (`line') and its attribute (`peak') in the chart.

While the task of ChartQA has received growing attentions in recent years, existing datasets have several major limitations: \Ni the questions are generated automatically using pre-defined templates \cite{figureqa,dvqa, leafqa,stlcqa} which lack naturalness,  \Nii the charts  are created automatically using a programming tool like Matplotlib \cite{stlcqa} which do not reflect the diverse styles of many real-world charts, and  finally, \Niii in most datasets, the answer comes from a small fixed sized vocabulary (\eg\ chart axis labels, `yes', `no'), ignoring many complex reasoning questions where the answer is derived through various mathematical operations such as aggregation and comparison.

Since most datasets only support \emph{fixed vocabulary} questions, existing 
models usually treat the task as a classification problem and rely on dynamic encoding techniques with the questions and answers encoded in terms of spatial positions of chart elements (\eg\  \textit{x-axis-label-1}).
Such approaches do not work when the OCR model generates errors or when the question refers to chart elements using synonyms 
(\eg\ US vs. United States).
PlotQA \cite{plotqa} attempts to support \emph{open vocabulary} questions by applying a TableQA model~\cite{sempre} but it does not consider any visual features of a chart which are critical for answering visual reasoning questions.

To address these limitations, 
we present a large-scale 
benchmark covering  9,608 human-written questions focusing on logical and visual reasoning questions. Since human annotations are costly, we 
also generated  another 23,111 questions automatically  from human-written chart summaries using a T5 model \cite{Raffel2020t5} 
and manually validated a subset of it for quality assurance.
In this way, we collect a large number of questions automatically while maintaining rich variations in language as they were generated from human-written summaries. Our benchmark consists of 20,882 charts which are curated from four different online
sources to ensure variety in visual styles and topics. 

To address the challenges introduced in our benchmark,
where many questions involve complex reasoning and visual references to charts, we propose an approach that combines visual features and extracted data from the chart image. Our pipeline first extracts the underlying data table from the chart image by adapting the ChartOCR model \cite{ChartOCR} as well as the visual features from the chart image using neural models. Then, we adapt two  transformer-based QA models where we utilize both the extracted data table and visual features of the chart in a unified way. Our models achieve the state-of-the-art results, or stands on par with the previous models on the previous datasets as well as on our newly created benchmark.

In sum, our main contributions are: \Ni A large-scale ChartQA dataset with real-world charts and human-authored question-answer pairs; \Nii a pipeline approach that combines visual features and automatically extracted data from charts to  utilize in transformer-based QA models that provide state-of-the-art results; and \Niii an extensive analysis and evaluation of the performance of our models. Our code and dataset are publicly available at
\url{https://github.com/vis-nlp/ChartQA}

\begin{table*}[t]
  \centering
  \small
  \scalebox{0.95}{
  \begin{tabular}{l|c|c|c|c|c}
  \toprule
    Datasets & \makecell{Question \\ Types} & \makecell{Answer \\ Types} &  \makecell{Real-world \\ Data} & \makecell{Real-world \\ Charts} & \makecell{
    \#Charts/\\ \#QA pairs}\\
    \midrule
    FigureQA \cite{figureqa} & Template-based & Fixed & \xmark & \xmark & 180K/2.3M\\
    DVQA \cite{dvqa} & Template-based & Fixed & \xmark & \xmark & 300K/3.4M \\
    LEAF-QA \cite{leafqa} & Template-based & Fixed & \cmark & \xmark & 240K/2M \\
    LEAFQA++ \cite{stlcqa} & Template-based & Fixed & \cmark & \xmark & 244K/2.5M \\
    PlotQA \cite{plotqa} & Template-based & Open & \cmark & \xmark & 224K/28M\\
    \midrule
    \textbf{ChartQA-H \blue{(ours)}} & \textbf{Human-authored} & \textbf{Open} & \cmark & \cmark & \textbf{4.8K/9.6K}\\
    \textbf{ChartQA-M \blue{(ours)}} & \textbf{Machine generated} & \textbf{Open} & \cmark & \cmark & \textbf{17.1K/23.1K}\\
    \bottomrule
  \end{tabular}}
  \caption{Comparison between existing  datasets and our new ChartQA benchmark}
  \label{tab:1}
\end{table*}

\section{Related Work}

\paragraph{Existing Datasets} ChartQA differs from previous datasets in two main aspects: the questions' types (human-authored vs. template-based) and the chart source (real-world vs. generated using a tool). A detailed comparison is shown in  \Cref{tab:1}.  Earlier datasets such as FigureQA \cite{figureqa}, DVQA \cite{dvqa}, LEAF-QA \cite{leafqa} and LEAF-QA++ \cite{stlcqa}  are mostly synthetic where the questions are generated using a small number of templates and the answers come from a fixed set of vocabulary (e.g. `yes', `no'). Moreover, their charts are created automatically using the same software. While FigureQA and DVQA use synthetically-generated data to plot the charts, LEAF-QA and LEAFQA++ use real-world data.  PlotQA \cite{plotqa} is the only dataset with open-vocabulary questions that require applying aggregation operations on the underlying chart data. However, they do not have visual reasoning questions while their questions are still template-based and the charts are plotted using a software. 
\citet{kim2020answering} ran a formative study with a very small human-authored dataset consisting of 52 charts and 629 QA pairs to understand how people ask questions about charts and explain answers. To our knowledge, there is no large-scale Chart QA dataset involving  visual and logical reasoning questions written by humans on real-worlds charts which motivated us to build a new dataset.

\paragraph{Existing Models} 
There are two main approaches for Chart QA. The first approach uses classification-based visual QA models that can only handle fixed-vocabulary questions \cite{leafqa, stlcqa, prefil, figureqa, dvqa}. These models use encoders to encode the question and the chart image and an attention mechanism to combine the features of both the question and chart before applying a classification layer. These models mostly utilize dynamic encoding techniques to encode the question in terms of the positional information of the textual elements in the chart image that are prone to OCR noise. The second approach applies table QA methods by either assuming that the data table of the chart is given \cite{kim2020answering, ahmed-workshop-2021} or by extracting it from the chart image using vision techniques \cite{plotqa}.

\paragraph{Chart Data Extraction}
Early papers introduced semi-automatic systems to extract the data from the chart images \cite{Revision, chartsense}. \citet{Choi2019VisualizingFT}, \citet{Liu2019DataEF}\change{, and \cite{Siegel2016FigureSeerPR} }proposed fully automatic chart data extraction pipelines, however, their methods rely on various heuristics which do not work for many real-world charts and the performance was still limited. \citet{ChartOCR} also automatically
extract data from real-world charts with high accuracy. Still, the model only predicts the raw data values of marks (\eg\ bars) without associating them with their corresponding axis or legends. We extend their pipeline to extract the fully-structured data table to pass it to our models.

\section{ChartQA Datasets}
\subsection{Data Collection \& Preparation}

To ensure that our 
benchmark
covers various topics and 
charts with a diverse range of styles, we crawled charts from four different sources: 
\Ni Statista ({\href{https://www.statista.com}{statista.com}}) is an online platform that presents charts covering a variety of topics including economy, politics, and industry.  
\Nii The Pew research {(\href{https://www.pewresearch.org}{pewresearch.org})} 
publishes report about social and economic issues, demographic trends and public opinion with a wide variety of charts. 
\Niii Our World In Data or OWID {(\href{https://ourworldindata.org}{ourworldindata.org})}  
is 
another platform that contains thousands of charts  about different global issues such as economy, finance, and society. 
\Niv Organisation for Economic Co-operation and Development or OECD {(\href{https://www.oecd.org}{oecd.org})} is a global organization which shares 
reports and data analysis for policymaking.

For the Pew dataset, we only crawled chart images since  the underlying data tables are not available. For the other three, we extracted the underlying data tables, metadata (\eg\ title, chart type), SVG file and associate text description.  Finally, we extracted the bounding boxes information of the different chart elements (\eg\ x-axis labels) from the SVG files to train our data extraction models.

\subsection{Data Annotation} \label{subsec:data-annot}
We have two main annotations procedures: \Ni collect human-authored QA pairs using Amazon Mechanical Turk (AMT) and \Nii generate QA pairs from the Statista human-written summaries. 

\paragraph{$\bullet$ Human-authored QA annotation} To create human-authored QA pairs, we designed an AMT task (see \ref{app:AMT} for details) in which we asked the crowdworkers to focus on two types of questions for each chart image: compositional and visual questions. Compositional questions contain at least two mathematical/logical operations like \emph{sum}, \emph{difference} and \emph{average}, while visual questions refer to the visual attributes such as \textit{color}, \textit{height}, and \textit{length} of graphical marks (\eg\ \textit{bars}) in the chart. We focus on these two types of questions because people tend to ask them commonly~\cite{kim2020answering, hoque-etal-2017} and previous datasets mostly do not focus on such complex visual and logical reasoning questions. 
For each chart, the workers provide two questions with the answers. The same questions are then answered
by another annotator. If both workers' answers exactly match, we consider the answer to be correct. Otherwise, we manually check the answers to select the final correct answer. 
Overall, the agreement between the crowd workers based on exact matches was 61.04\%. 
\change{However, such exact match does not consider typos or lexical variations  (e.g., 3\$ vs. 3 dollars, 86.33 vs 86.3) that are common in human annotation. Hence, we have also manually checked the agreement on 500 random samples and found the agreement to be much higher (78.55\%) when we consider typos and lexical variations.}

\paragraph{$\bullet$ Dataset Augmentation}

Prior work on QA  has performed data augmentation by either creating template-based  or machine generated questions, \eg\ for visual QA \cite{visual-data-augmentation}  and textual QA \cite{paq}. Template-based questions generally lack rich linguistic variations. On the other hand, large-scale language models like T5 \cite{Raffel2020t5} which are trained on very large data from various web sources can learn general linguistic properties and variations \cite{NEURIPS2020_1457c0d6}. Therefore, we opt for the latter.

Specifically, we fine-tune a pre-trained T5 model on the SQuAD QA dataset  \cite{squadv1} and apply to the human-written chart summaries that come with the charts from Statista to automatically generate questions that are human-like with sufficient lexical and  syntactic variations. The process involves training and applying two T5 models: one for \emph{answer extraction} and the other for answer-aware \emph{question generation}. For answer extraction, the T5 model is trained to generate possible answers separated by \texttt{[SEP]} token given the textual summary as input (\ie\ trained on SQuAD's \emph{passage} $\rightarrow$ \emph{answer} pairs). For question generation, the proposed answer is first concatenated with the summary in the  format: \texttt{Answer}: \textit{Answer} \texttt{Context:} \textit{Chart Summary}.  Then, the T5 model is trained to generate a question from the given question using the chart summary. This model is trained on SQuAD's \emph{(passage, answer)} $\rightarrow$ \emph{question} pairs. Since the summaries are human-written, the generated questions are similar to the human-authored questions (see example questions in \ref{app:examples}).

\begin{table}[t!]
\centering
\small 
\scalebox{0.9}{
\begin{tabular}{lcc|cc}
\toprule
\multirow{2}{*}{Split} & \multicolumn{2}{c}{\textbf{ChartQA-H}} & \multicolumn{2}{c}{\textbf{ChartQA-M}} \\
 & Charts & Questions & Charts & Questions \\
\midrule
Training &     3,699 &  7,398 & 15,474 & 20,901\\
Validation & 480   &  960   & 680 & 960\\
Test & 625 &   1,250 & 987  & 1,250\\
\midrule
Total &  4,804  & 9,608 & 17,141 & 23,111\\
\bottomrule
\end{tabular}
}
\caption{\small Our dataset statistics for each split.
}
\label{tab:splits}
\end{table}

However, the T5 question generation model may still generate invalid questions because of the mismatch in training and test domains. We notice that some questions are either incomplete or not answerable from the chart (\eg\ `What province includes Cape Town?' is not answerable because it requires knowledge outside of the chart).
To filter out such invalid questions, we developed a simple heuristic where we filter out the question if the answer cannot be found in the chart data table. This heuristic was inspired by the fact that most answers to the generated questions were  values/labels of chart elements.
After applying the heuristic, we manually analyzed 1,250 QA pairs and found that {86.64\%} of them were \change{complete, answerable, and correct} given the chart. Moreover, for the sake of fair  evaluation, we manually cleaned the test set of the machine generated dataset by removing invalid questions.

\paragraph{$\bullet$ Data split}
We randomly split both of the human-written (ChartQA-H) and machine generated (ChartQA-M) QA pairs into train, validation, and test sets as shown in  \Cref{tab:splits}.

\begin{table}[t!]
\centering
\small
\scalebox{0.90}{\begin{tabular}{lcccc|l}
\toprule
Type & Statista-H & Pew & OWID & OECD & Statista-M \\
\midrule
Bar &  1,696    & 783    &   507   &  128 &  15,223   \\
Line &  401       & 249    & 279     &    103 &  1,768  \\
Pie &    387     &  271   &    0  &   0 &  150\\
\midrule
Total &   2,484 &   1,303 & 786  &  231 &  17,141 
\\
\bottomrule
\end{tabular}
}
\caption{\small Number of charts from each source. Statista-H and Statista-M refer to the datasets with human-written and machine generated questions respectively from Statista
}
\label{tab:dataset-chart-count}
\end{table}

\begin{table}[t!]
\centering
\small 
\scalebox{0.95}{
\begin{tabular}{p{1.9cm} p{4.5cm} p{0.2cm}}
\toprule
\small{\textbf{Type}} & \small{\textbf{Example}} & \small{\textbf{\%}}\\
\midrule
\small{\emph{Data retrieval} } & \small{What's the percentage of men who thinks Valentine's Day is overrated?} & \small{13.0}   \\ \\[-1em]
\small{\emph{Visual} } & \small{What is the value of the \blue{rightmost light blue bar}?} & \small{10.7} \\ \\ [-1em]
\small{\emph{Compositional } } &  \small{
How many years does the poverty percentage rose above 11\%?
} & \small{\textbf{43.0}}\\ \\[-1em]
\hspace{-0.3em}{\small{ \emph{Both visual \& compositional}}} & \small{Between \blue{the second and the third} age groups \blue{from the left}, which opinion deviates the most?}   &  \small{33.3}        \\
\bottomrule
\end{tabular}
}
\caption{\small Distribution of  questions types of among 300 randomly chosen human written questions (blue-colored tokens make visual references to the chart).
}
\label{dataset-content}
\end{table}

\subsection{Dataset Analysis}

Our dataset has three commonly used chart types: bar, line, and pie charts (\Cref{tab:dataset-chart-count}). Bar is the most common type of chart across all datasets as they are quite prevalent in real-world sources. We further categorize the bar and line charts into simple vs complex where  data tables of simple charts have only two columns where complex charts involve multiple columns (\eg\ stacked or grouped bars and multi-line charts).
Among bar charts, 79.4\% were simple and 29.6\% were complex. For line charts, 61.0\% were simple and 39.0\% were complex.

\begin{figure*}[t!]
  \includegraphics[width=1\textwidth]{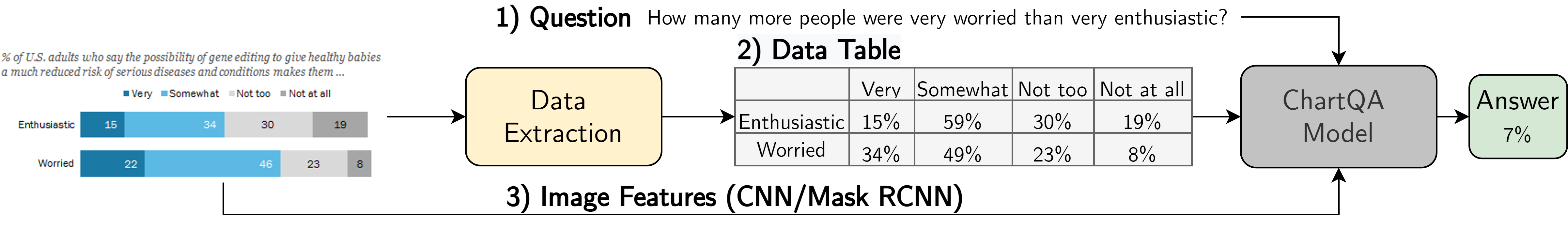}
  \caption{\small Our approach for question answering over charts. 
  If not provided,
  the underlying data table is first extracted from the chart image 
  using ChartOCR.
   We then pass the extracted data table in addition to the question and the image features to the ChartQA model where the ChartQA model represents one of the following: TaPas, VisionTaPas, T5, and VL-T5.}
  \label{Pipeline}
\end{figure*}

We have also analyzed the basic linguistic statistics about our benchmark (see \ref{app:dataset-analysis}). Unlike previous datasets, our benchmark has more unique tokens on both types of QA pairs and on both questions and answers -- 6,150 and 4,319 unique tokens in questions and answers respectively in ChartQA-H whereas 12,379  and 11,979 unique tokens in questions and answers respectively in ChartQA-M. We also observe that questions cover a variety of syntactic structure and sometimes exhibit informal languages and typos. 
Overall, this suggests the richness of language variations which may introduce more challenges to the task.
Finally, the topic distribution in our data is quite diverse as it is  constructed from four different sources. Politics is a common topic among all sources but particularly in the Pew dataset where nearly half of charts  are about U.S. Politics \& Policy (45.4 \%).
Other common topics include economy, health, and society.

 To analyze the nature of questions, we randomly selected 300 QA pairs from our benchmark and categorized them into four types (\Cref{dataset-content}). We see that the vast majority of questions (76.33\% in total) are either compositional or both visual and compositional, which reflects the real-world scenarios where people ask complex reasoning questions. We also find that people make visual references to a variety of visual attributes of marks (see \ref{app:dataset-analysis}), most commonly to \textit{color} (\eg\ `orange line') and \textit{length} (\eg\ `tallest bar') followed by  \textit{size} (\eg\ `largest  slice') and \textit{position} (\eg\ `leftmost bar'). 


\section{Method}

\subsection{Problem Formulation \& Data Extraction}
\label{sec:problem-extraction}
The overall process of our ChartQA system is shown in \Cref{Pipeline}. We consider two problem settings for ChartQA. The first setting assumes that the underlying data table of the chart image is available. Formally, we are given a dataset with $N$ examples $\gD = \{c_i, t_i, q_i, a_i\}_{i=1}^N$, where $c_i$ represents a chart image, $t_i$ represents the underlying data table, $q_i$ represents a question over $c_i$, and $a_i$ represents the answer to the question. The ChartQA models learn to predict the answer $a_i$ given $c_i$, $t_i$ and $q_i$. 

The gold data tables are not generally accessible in most real-world scenarios. Thus we consider the second setup where the underlying data table $t_i$ for chart image $c_i$ is extracted by adapting a state-of-the-art ChartOCR \cite{ChartOCR}. ChartOCR first  locates the main elements of the chart image (\eg\ plot area, title) as well as data-encoding marks (\eg\ bars
) using key-point detection networks. It then uses the detected keypoints of each mark along with  axis-labels to estimate the data value of that mark. However, 
it does not associate the predicted data values with corresponding text labels (\eg\ x-axis-label).
 Hence, we extend their approach to output the fully-structured data tables. We utilize the CRAFT \cite{craftrecognition} model to recognize the texts in the chart elements. Then, we associate the data values with their text labels using positional and color information (see \ref{app:data_extraction} for details).

\subsection{Models} \label{subsec:data2text}

Our approach to ChartQA builds on two of the state-of-the-art TableQA models: T5 \cite{Raffel2020t5,nan2021feta} and {\sc{TaPas}} \cite{tapas}. The input to these models consists of the question $q_i$ and the data table $t_i$. Different from TableQA, ChartQA often involves extracting visual information from chart images. For this, we also experiment with the visual counterparts of the TableQA models that also take the chart image features into account. While T5 has a visual variant, VL-T5 \cite{vlt5}, {\sc{TaPas}} does not. In this work, we extend Tapas to consider the image features and call it Vision{\sc{TaPas}}. More details on models 
are provided in \ref{app:baselines}.

\begin{figure*}[t!]
\begin{subfigure}[b]{.3\textwidth}
\centering
    \includegraphics[width=1\textwidth,keepaspectratio]{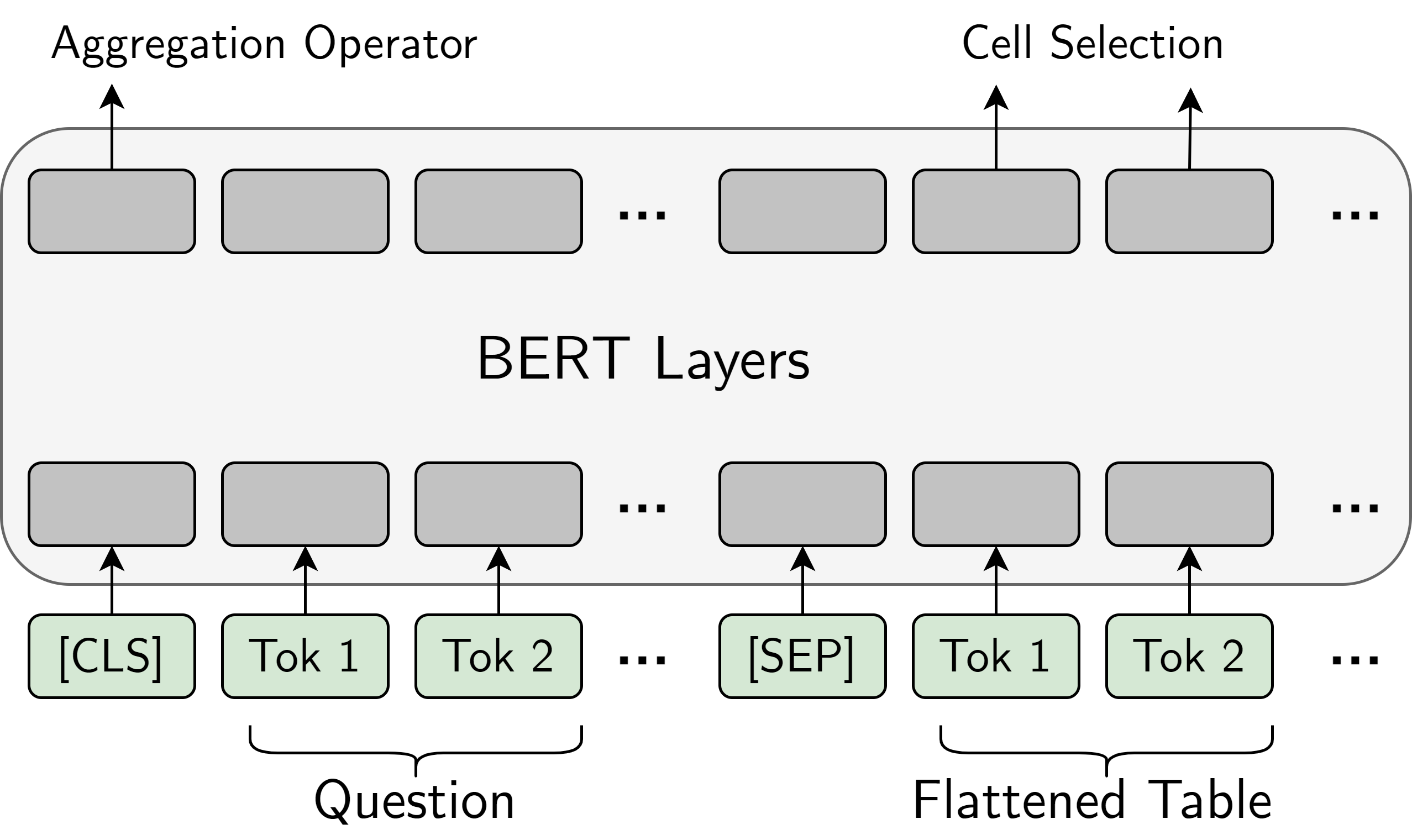}
\caption{\small {\sc{TaPas}}}
\label{fig:tapas}
\end{subfigure}
\begin{subfigure}[b]{.68\textwidth}
\centering
    \includegraphics[width=1\textwidth,keepaspectratio]{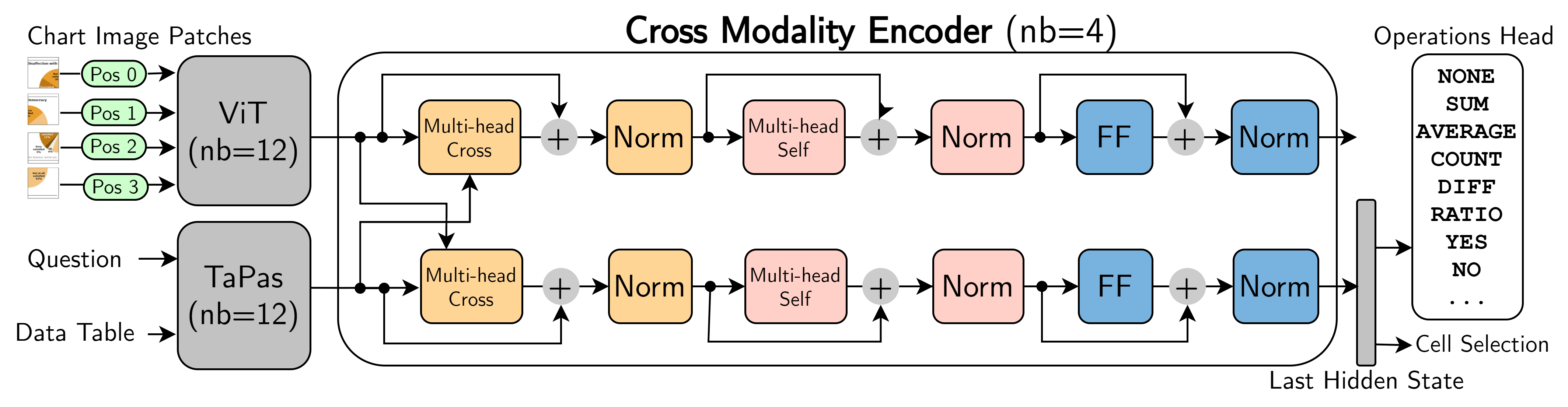}

\caption{\small {\sc{VisionTaPas}}}
\label{VisionTapas}
\end{subfigure}
\caption{\small  TaPas and VisionTaPas models. TaPas adds positional embeddings to the tokens to encode the tabular structure of the data table. VisionTaPas uses a cross-modality encoder  to combine visual features from ViT and outputs from TaPas encoders.}

\label{models}
\end{figure*}

\vspace{0.2em}
\noindent \textbf{$\bullet$ {\sc{T5}}} \cite{Raffel2020t5} is an encoder-decoder model which unifies the NLP tasks as text-to-text generation using the same architecture and loss function. It has been pre-trained on massive amount of unlabelled data with a self-supervised denoising objective. To fine-tune T5 on our ChartQA task, we flatten the data table and feed it along with the question as: "\texttt{Question:} \textit{Question {tokens}} \texttt{Table:} \textit{Flattened table {tokens}}", and the model is trained to generate the answer directly. 

\vspace{0.2em}
\noindent \textbf{$\bullet$ {\sc{VL-T5}}} \cite{vlt5} is an extension of T5 that unifies the Vision-Language (VL) tasks as text generation {conditioned on multimodal inputs}. The input consists of both textual tokens and visual features of the objects extracted from the image using Faster R-CNN \cite{RenHG015}. The model is pre-trained on multiple multimodal tasks such as language modeling, visual QA, and visual grounding. 
We utilize VL-T5 for our ChartQA task in the following manner. For the textual input, we do the same as T5 where we flatten the data table of the chart image and concatenate it with the question text. For the visual input, we extract the visual features of different marks in the chart image (\eg\ bars, lines) using Mask R-CNN \cite{maskrcnn} with Resnet-101 as its backbone (see \ref{app:maskrcnn} for details). Unlike the original VL-T5 where a fixed number of objects is provided (36), the number of elements varies from one chart to another. To account for this, we pad the extracted visual features with zeros to have a fixed length of 36.

\vspace{0.2em}
\noindent \textbf{$\bullet$ {\sc{TaPas}}} \cite{tapas} extends a BERT \cite{BERT} architecture with additional positional embeddings for rows and columns to encode a table. 
As shown in \Cref{fig:tapas}, the input to the model has the following format: \texttt{[CLS]} \textit{Question tokens} \texttt{[SEP]} \textit{Flattened table tokens}. The tokens are encoded with the table-specific positional embeddings in addition to BERT's segment and positional 
embeddings. The model has two output heads: aggregation operation head and cell selection head. The aggregation operation head predicts an operation (e.g., {\sc{count}}, {\sc{sum}}, {\sc{average}}, {\sc{none}}) which is then applied to the cell values selected  by the cell selection head. Depending on the operation type, the selected cells can constitute the final
answer or the input used to infer the final answer. 

TaPas is first pre-trained on masked language modeling objective using table-text pairs crawled from Wikipedia where table cells are randomly masked and the model is trained to predict them. It is then fine-tuned in a weakly-supervised  manner (using answers as the only supervision) with end-to-end differentiable objectives.

\vspace{0.2em}
\noindent \textbf{ $\bullet$ VisionTaPas} is our extension of TaPas for QA over charts. It consists of three main components: a vision transformer encoder for encoding the chart image, a TaPas encoder for encoding the question and data table and a cross-modal encoder (\Cref{VisionTapas}).

\textbf{Vision Transformer} or ViT \cite{vit} utilizes the  transformer encoder architecture \cite{VaswaniSPUJGKP17} in vision tasks. Given a 2D chart image, the image is divided into a sequence of 2D patches $\{\vp_1, \ldots, \vp_n\}$. Each patch is then flattened and linearly projected into a $d$-dimensional embedding vector. To incorporate the positional information of the patches, 1D learnable positional embeddings are added to the image features. An $L$-layer ViT encoder produces a sequence of embeddings $\mH =  \{\vh_{\text{cls}}^L, \vh_1^L, \ldots, \vh_n^L\}$ representing the special \texttt{[CLS]} token and the image patches. We initialize the ViT module with the pre-trained weights from \cite{vit}.   

The \textbf{TaPas} encoder is utilized in the same manner as described above to encode the tokens in the question and the data table. For an input token sequence $\{w_{\text{cls}},  w_1, \ldots, w_m\}$, an $L$-layer TaPas generates the corresponding encodings $\mZ = \{\vz_{\text{cls}}^L, \vz_1^L, \ldots, \vz_m^L\}$. This module is initialized with the TaPas weights \cite{tapas} pre-trained on the WikiTQ dataset \cite{sempre}. 

The \textbf{Cross-modality Encoder} 
takes the output of ViT and TaPas encoders ($\mH$ and $\mZ$) and compute multimodal encodings. It has four blocks, each containing 
a visual branch and a textual-tabular branch. The input first passes through the multi-headed cross attention layers in parallel, where in the visual branch the query vectors are the visual features, and the key and context vectors are the textual-tabular features and vice versa in the textual-tabular branch. The cross-attended features are then passed through a self-attention layer followed by a fully connected layer. Similar to the transformer model, each layer applies layer normalization \cite{LayerN} and is wrapped with a residual connection. Finally, we append the aggregation operation and the cell selection heads of TaPas to the final layer at the textual-tabular branch.

\begin{table*}[t!]
    \centering
    \scalebox{0.66}{\begin{tabular}{l|cccccccc|cc}
        \toprule
       \textbf{Models} & \multicolumn{4}{c}{\textbf{FigureQA}} & \multicolumn{2}{c}{\textbf{DVQA} (ORACLE / OCR)} & \multicolumn{2}{c}{\textbf{PlotQA}} & \multicolumn{2}{c}{\textbf{ChartQA}}\\

         & Val1 & Val2 & Test1 & Test2 & Test-Familiar & Test-Novel \cmmnt{&  Val}  & Test V1 & Test V2 & Val & Test \\
        \midrule
        \multicolumn{10}{c}{\textbf{Gold Data Table Provided}}  \\
         {TaPas} & \textbf{98.10\%} & \textbf{98.09\%} & - & - & 53.40\% & 53.40\%  & \cmmnt{21.53\% &}  21.56\% & 19.55\% & 49.16\% & 51.80\% \\
       {VisionTaPas} &  97.59\% &  97.96\% & - & - & \textbf{99.36\%} & \textbf{99.37\%} \cmmnt{&  80.17\%} & 80.18\% & 58.29\% & \textbf{59.32\%} & \textbf{61.84\%} \\
        {T5}& 95.75\% & 95.75\% & - & - & 94.33\% & 81.42\% \cmmnt{& 93.25\%} & 93.24\% & \textbf{85.99\%} & 59.11\% & 59.80\% \\
       {VL-T5} & 96.45\% & 96.43\% & - & - & 98.90\% & 80.18\% & \cmmnt{\textbf{96.35\%} &} \textbf{96.38\%} & 84.70\% & 58.80\% & 59.12\% \\
        \midrule
        \multicolumn{10}{c}{\textbf{Gold Data Table Not Provided}} \\
\rowcolor{gray!20}        {TaPas} &  90.32\% & 90.43\% & 89.52\% & 89.57\% & 50.28\% / 48.82\% & 50.24\% / 48.68\% \cmmnt{& 15.12\%} & 15.09\% & 12.90\% & 39.68\% & 41.28\% \\
\rowcolor{gray!20}        {VisionTaPas} & 91.46\% & 91.45\% & 90.68\% & 90.64\% & 95.38\% / \textbf{94.43\%} & 95.46\% / \textbf{94.54\%} & \cmmnt{65.28\% &} 65.30\% & 42.50\% & \textbf{42.60\%}& \textbf{45.52\%} \\
\rowcolor{gray!20}        {T5}& 87.97\% & 87.83\% & 87.56\% & 87.57\% & 90.20\% / 89.01\% & 77.97\% / 76.89\% & \cmmnt{72.67\% &} 72.62\% & \textbf{56.22\%} & 40.15\% & 41.04\% \\
\rowcolor{gray!20}        {VL-T5} & 88.60\%  & 88.49\% & 88.20\% & 88.18\% &  94.80\% / 93.75\% & 77.04\% / 76.14\% & \cmmnt{\textbf{75.92\%} &} \textbf{75.90\%} & 56.02\% & 38.43\% & 41.56\%\\
\rowcolor{gray!20}        {PReFIL} & \textbf{94.84\%} & \textbf{93.26\%} & \textbf{94.88\%} & \textbf{93.16\%} &  96.37\% / 80.88\% & 96.53\% / 80.04\% \cmmnt{& -} & - & - & 4.53\% & 4.8\%\\
\rowcolor{gray!20}        {PlotQA*} & - & - & - & - & --------- / 57.99\%  &  ---------  / 59.54\% \cmmnt{& -} & - & 22.52\% & 36.15\% & 38.00\%  \\
\rowcolor{gray!20}        {STL-CQA} & - & - & - & - & \textbf{97.35\%} /  ---------  & \textbf{97.51\%} /  ---------  & - & - & - \cmmnt{& -}& - \\
        \bottomrule
    \end{tabular}}
    \caption{\small Evaluation results for different models. 
    For DVQA, we have reported the results with and without using Oracle for OCR. We do not evaluate on FigureQA test sets with the gold data table setup since they do not have ground data tables.
    }
\label{tab:evaluation-table}
\end{table*}


\vspace{0.5em}
\noindent \textbf{Extension to Other Operations} Many questions in our ChartQA dataset require performing a subtraction or ratio operation, which the original TaPas model does not support. We thus extend the operation head to add those two operations (\Cref{VisionTapas}). However, instead of training them in a weakly-supervised manner based on the final answer (as done in TaPas), we find it more effective when provided with more direct but potentially noisy supervision on the cells to consider. We rely on some heuristics to generate such supervision in our training data. For example, given a question ``What's the difference between A and B?'', an answer $5$, and data values ``3, 6, 8'', we look for two values between which the difference is 5 (i.e. 8 and 3). While this may yield noisy supervision, similar approaches have been successfully exploited to inject reasoning capability in neural models \cite{geva-etal-2020-injecting,saxton2018analysing}; on a random sample of 100 such questions, a manual checking shows 24\% noise  with our heuristics. To handle the fixed vocabulary answers (e.g. `Yes', `No'), we further extend the operation head to include those classes.

\section{Evaluation}

\subsection{Datasets, Baselines \& Metrics} 

We evaluate our models on three datasets from previous work namely, FigureQA \cite{figureqa}, PlotQA \cite{plotqa} and DVQA \cite{dvqa}, as well as our newly created ChartQA dataset. We compare our benchmarking models (\Cref{subsec:data2text}) with two following baselines\footnote{Two other datasets (LeafQA, LeafQA++) and baselines  (STL-CQA, LEAF-NET) are not publicly available}:

\vspace{0.2em}
\noindent \textbf{$\bullet$ {\sc{PReFIL}}} \cite{prefil} is a classification approach that fuses the question and image features in parallel. The features are then aggregated and projected into a final classification layer.

 \vspace{0.2em}
\noindent \textbf{$\bullet$ {\sc{PlotQA*}}} is our reimplementation of PlotQA~\cite{plotqa}. It 
parses the chart image to extract the underlying data table and then employs a TableQA model  
from \citet{sempre}. However, since their data extraction approach is specific to their synthetic dataset that does not generalize well to real-world charts, 
we use data tables extracted according to our method  (\cref{sec:problem-extraction}) to evaluate their approach. 

Following \citet{plotqa}, we use a relaxed accuracy measure for the numeric answers to allow a minor inaccuracy that may result from the automatic data extraction process.
We consider an answer to be correct if it is within 5\% of the gold answer. For non-numeric answers, we still need an exact match to consider an answer to be correct.

\subsection{Results}

\noindent \textbf{Previous Datasets~~} 
When the gold data table is provided, VisionTaPas and VL-T5 achieve near perfect results, however, the performance slightly decreases when it is not provided 
(\Cref{tab:evaluation-table}).
Still, VisionTaPas and VL-T5 achieve state-of-the-art results on DVQA (fully-automated setup) and PlotQA \change{V1} datasets, respectively. For example, VisionTaPas achieves 94.54\% accuracy in the DVQA test set (14.5\% margin over PReFIL). Moreover, our approach proved to be more robust to OCR noise. 
Unlike PReFIL whose performance significantly dropped by 16.49\% when using OCR outputs
instead of ORACLE, VisionTaPas only witnessed a marginal decrease in performance (0.92\%). Similarly, in the PlotQA dataset, both models have outperformed the PlotQA model by wide margins. 
Another observation is that the improvement of VL-T5 over T5 is limited only to the PlotQA \change{V1} dataset  likely due to the lack of visual reasoning questions.  \change{In fact, the performance of both models is quite similar on PlotQA V2 test set where the majority of the questions are not visual.} Finally, 
while the TaPas model achieves the best results on FigureQA (Gold Table setup), it does not perform very well on DVQA and PlotQA. 
This is likely because most questions in FigureQA are answerable from the data table alone. In PlotQA, however, questions are not always answerable from the data table alone and may involve the difference and ratio operations which are not supported by TaPas. This highlights the importance of the extensions we have made in the VisionTaPas model.

\noindent \textbf{ChartQA Dataset~~}  We observe that VisionTaPas achieves state-of-the-art performance on both problem scenarios. PReFIL performs pooly (4.8\%) as it is a classification model which does not work well for  the open-vocabulary questions in our dataset. We also notice VL-T5 does not necessarily improve over T5, likely because many visual questions in our new dataset involve 
multiple references to chart elements and VL-T5 cannot effectively capture such references.
Overall, 
the accuracies of different models are generally lower in our dataset compared to previous datasets, suggesting the challenges introduced with the human-written 
visual and logical reasoning questions. Finally, the performance of our models 
decreases when the gold data table was not given.
This highlights the increasing challenge of automatic data extraction from real-world charts with diversity in 
styles.

We also evaluate the \emph{transferability} of the models and the datasets, where we first pretrain the two top performing models (VisionTaPas and VL-T5) on the PlotQA dataset and then fine-tune them on ChartQA. From 
\Cref{tab:pretrained}, we notice that the accuracy increased from 41.56\% to 51.84\% for VL-T5 while the improvement for VisionTaPas was marginal (1.56\%). One possible explanation is that VisionTaPas does not support nested arithmetic operations which are prevalent in ChartQA, so pretraining does not have a substantial effect. In contrast,  
we observe that the performance gain for VL-T5 were mainly for the compositional questions that do not require nested operations.
Overall, this suggests that large
datasets like PlotQA can be 
useful for pretraining the model even if the questions are generated from a small number of templates. 

\change{We also performed an another experiment in which we train the VL-T5 and VisionTaPas on the PlotQA dataset and evaluate directly on the ChartQA dataset without any fine-tuning. 
As shown in Table \ref{tab:pretrained}, the performance of the models decreased by wide margins when they are trained on the PlotQA dataset instead of the target dataset (e.g,. 45.52\% to 31.96\% for VisionTaPas). This supports our hypothesis that our newly created dataset, ChartQA, introduces more challenging visual and compositional questions and more lexical variations which the previous datasets lack. }

\subsection{Ablation Studies}
To assess the importance of extensions we made in the VisionTaPas model, we conducted an ablation study in which we remove the supervision for `difference' and `ratio' operations  from the 
model. 
The overall accuracy dropped 
by 1.80\%
and the accuracy on ChartQA-H (which have many such questions) dropped 
by 4.76\%
which suggests the usefulness of these operations (Table \ref{tab:pretrained}).

\begin{table}[h!]
\centering
\scalebox{0.68}{\begin{tabular}{lccc}
\toprule
Model & \textbf{ChartQA-H} & \textbf{ChartQA-M} & \textbf{Overall} \\
\hline
TaPas & 28.72\% & 53.84\% & 41.28\% \\
VisionTaPas & \textbf{29.60\%} & 61.44\% & \textbf{45.52\%} \\
VisionTaPas\dagger & 24.84\% & \textbf{61.60\%} & 43.72\% \\
T5 & 25.12\% & 56.96\% & 41.04\% \\
VL-T5 & 26.24\% & 56.88\% & 41.56\% \\
\midrule
VisionTaPas$^\star$ & 25.12\% & 38.80\% & 31.96\% \\
VL-T5$^\star$ & 22.08\% & 19.84\% & 20.96\% \\

\midrule
\rowcolor{gray!20} VisionTaPas \textbf{Pretrained} & 32.56\% & 61.60\% & 47.08\%\\
\rowcolor{gray!20} VL-T5  \textbf{Pretrained} & \textbf{40.08\%} & \textbf{63.60\%} & \textbf{51.84\%} \\
\bottomrule
\end{tabular}}
\caption{
    \small Accuracy of the different models on our benchmark.
    VisionTaPas$\dagger$ does not support difference and ratio operations. \change{VisionTaPas$^\star$ and VL-T5$^\star$ are trained on PlotQA and evaluated directly on ChartQA. }
}
\label{tab:pretrained}
\end{table}

We further analyze 
the performance by
chart types and question types (see \ref{app:results}). 
VisionTapas and VL-T5 perform better on bar charts while the performance decreases for other charts mainly due to higher data extraction errors, especially for pie charts which are less common in our dataset. To analyze question types, we randomly sampled 200 
human-written questions.
 As expected, the performance is much higher on the data retrieval questions that do not require mathematical reasoning  while the performance is lower for visual questions which refers to chart elements.

\subsection{Qualitative Analysis}

\begin{figure}[t!]
  \includegraphics[width=0.48\textwidth]{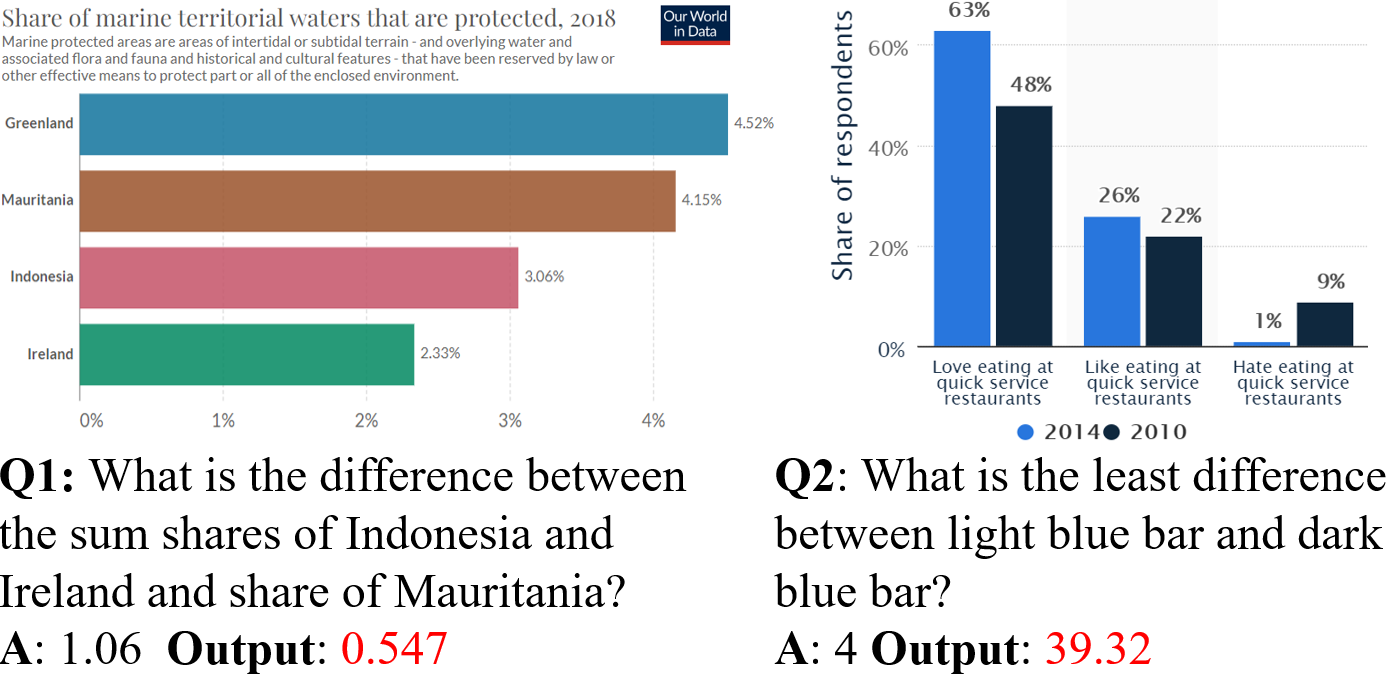}
  \caption{Example of errors from VisionTaPas
  }
  \label{results}
\end{figure}
We have manually analyzed model predictions to investigate the key challenges existing models face (see sample predictions in \ref{app:examples}).

\vspace{0.2em}
\noindent \textbf{Logical Inference with Nested Operations}
While VisionTaPas and VL-T5 handle various  mathematical/logical operations, still they cannot effectively handle nested operations.  For example, \textit{Q1} in \cref{results} requires the model to   add two numbers and then subtract from another number, but our model only outputs the difference between two numbers.In  future, we will extend the VisionTaPas model (by possibly training it in a sequential fashion \cite{Cho2018AdversarialTA}) to address the issue.

\vspace{0.2em}
\noindent \textbf{Input Representation} Complex visual compositional questions 
may require a multi-stage reasoning process (\eg\ \textit{Q2} in \cref{results}). Currently, our models take the data table and the visual features of the chart separately and then
combine them. Such 
representation does not fully capture the chart structure.
In future, we will develop better representations including semantic graph representations \cite{Teney2017GraphStructuredRF} that can 
exploit the 
relations among the question, chart objects, and data values.

\vspace{0.2em}
\noindent \textbf{Computer Vision Challenges} Table \ref{tab:evaluation-table} indicates that performance of our models decrease 
when the gold table is not given,
 suggesting the need for more accurate data extraction. Current approaches for  automatic data extraction are modular and combine deep learning and rule-based methods which are error-prone. An end-to-end deep learning approach could help improve the performance and generalize well to different chart styles.

\section{Conclusion}
We present ChartQA, a new large-scale benchmark with human-written questions focusing on visual and logical reasoning. We also introduce a new approach that combines visual features and extracted data table from a chart to answer questions. While our evaluation highlights the promise of this approach, it also reveals several unique challenges emerge from the  visual and logical reasoning questions asked by human which exhibit the informal, intricate, and nuanced nature of language. We hope that our benchmark  will serve as a starting point for others to address these challenges.

\section*{Acknowledgement}
The authors would like to thank the anonymous reviewers for their helpful comments. This research was supported by the Natural Sciences \& Engineering Research Council (NSERC) of Canada.

\section*{Ethical Considerations}

During the dataset collection and annotation process, we have considered several ethical issues. To respect the intellectual property of dataset sources, we only used  the publicly available charts that comply with their terms and conditions. According to Statista publication rights,\footnote{\href{https://www.statista.com/getting-started/publishing-statista-content-terms-of-use-and-publication-rights}{https://www.statista.com/getting-started/publishing-statista-content-terms-of-use-and-publication-rights}} users are given open access to the publicly available charts for academic purposes. According to the terms and conditions for Pew,\footnote{\href{https://www.pewresearch.org/about/terms-and-conditions/}{https://www.pewresearch.org/about/terms-and-conditions/}} users are allowed to download and publish the content as long as they are attributed to the Center or are not attributed to a different party. 
According to OECD \footnote{\href{https://www.oecd.org/termsandconditions/}{https://www.oecd.org/termsandconditions/}} terms and conditions, users can crawl and use the data in their own work for any purpose unless where restrictions apply. According to OWID \footnote{\href{https://ourworldindata.org/faqs\#can-i-use-or-reproduce-your-data}{https://ourworldindata.org/faqs\#can-i-use-or-reproduce-your-data}} terms and conditions, all their data are open access and users can download or utilize the data in their own work.

In order to fairly compensate the Mechanical Turk annotators, we considered the minimum wage in the United States at the time (\$7.25 USD per hour). The estimated time taken for each task is 3-5 minutes. Hence, these annotators received \$0.6 USD for each task. Additionally, to protect the privacy of these annotators, all of their annotations were anonymized.

To ensure the reproducibility of our experimental results, our hyperparameters settings are provided in \Cref{app:baselines}.

Our models can be abused to mislead the public about the charts content and implications. While our models provide state-of-the-art results on most of the existing datasets, we can not guarantee that their output will be correct all the time.

\bibliographystyle{acl_natbib}
\bibliography{chart2text}
\newpage
\appendix
\section{Appendices}
\subsection{Additional Details on Data Annotation}
\label{app:AMT}

\begin{figure*}
    \centering
    \includegraphics[width=0.85\textwidth]{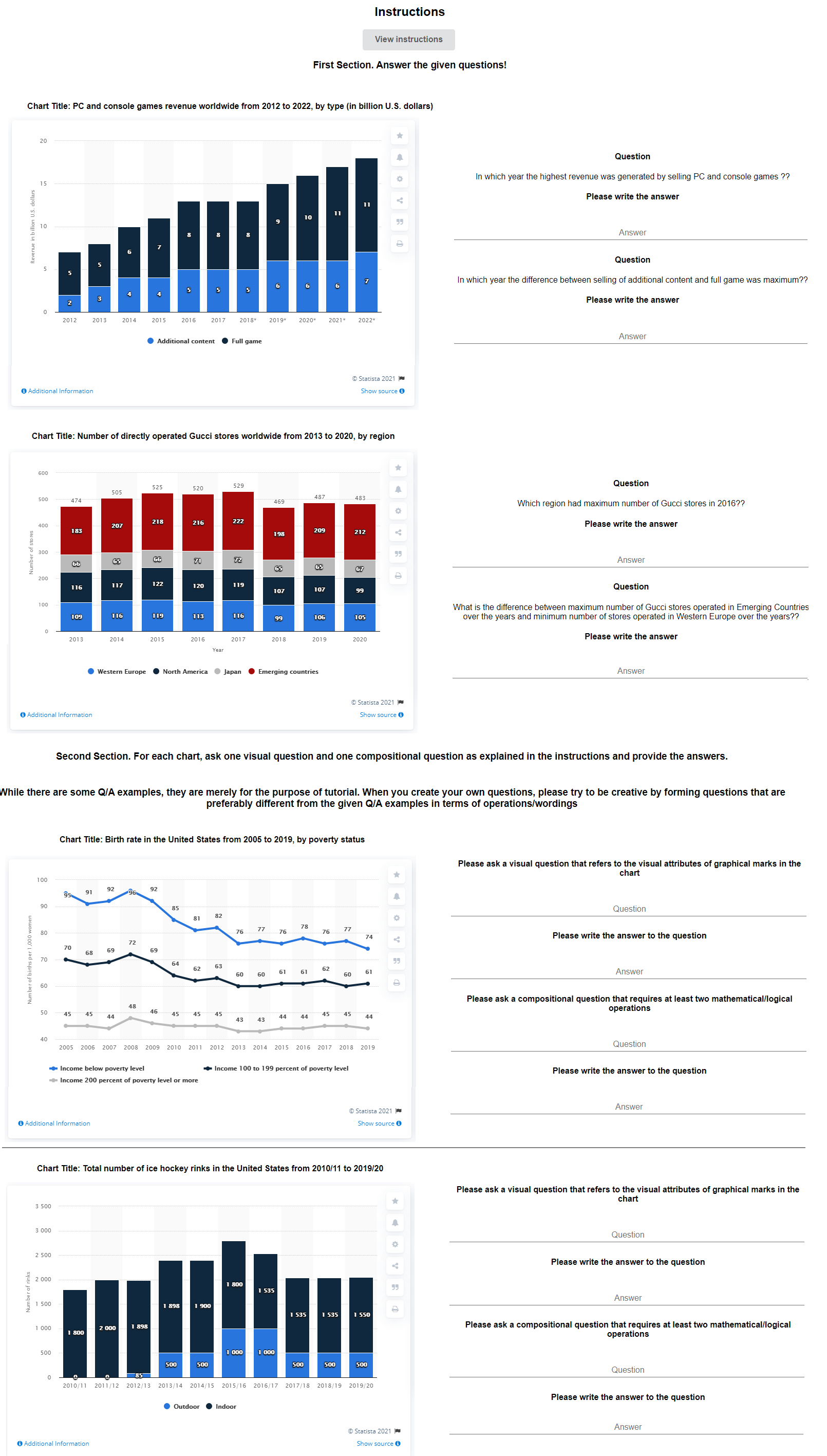}
    \caption{The user interface for the annotation task}
    \label{annot-interface}
\end{figure*}

\textbf{Amazon Mechanical Turk Task:} In each HIT (Human Intelligent Task), the workers verify two previously asked questions by other workers and also provide two new QA pairs. To ensure quality, we selected workers with an acceptance rate of 95\% and total accomplished HITs of 5000. Moreover, we further filtered the workers by giving them a pre-test to select the best qualified workers for this task. The data collection interface is shown in Figure \ref{annot-interface}. While presenting the chart, we ensure that the data labels of chart elements are visible to workers so that they can accurately perform the necessary arithmetic and logical operations to provide and answer the questions successfully.

\subsection{Dataset Analysis}
\label{app:dataset-analysis}
\begin{figure*}[t!]
\begin{subfigure}[b]{.49\textwidth}
\centering
    \includegraphics[width=\textwidth]{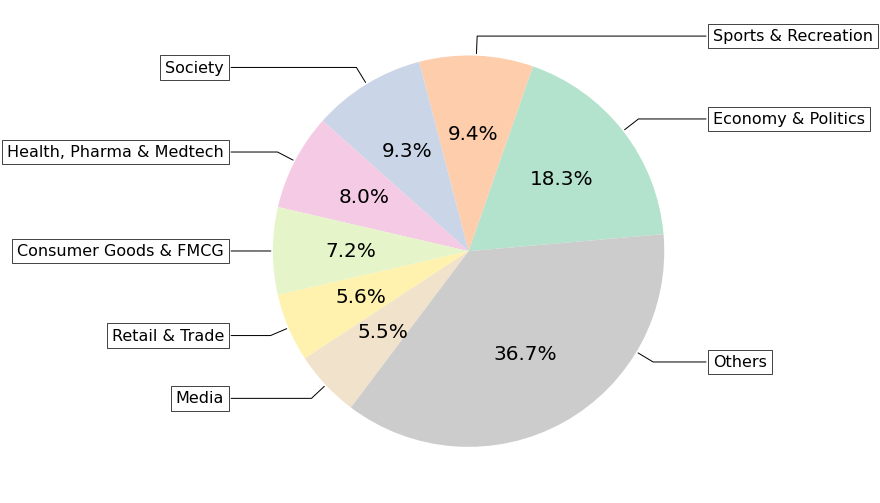}
\caption{\small Statista}
\end{subfigure}
\begin{subfigure}[b]{.49\textwidth}
\centering
    \includegraphics[width=\textwidth]{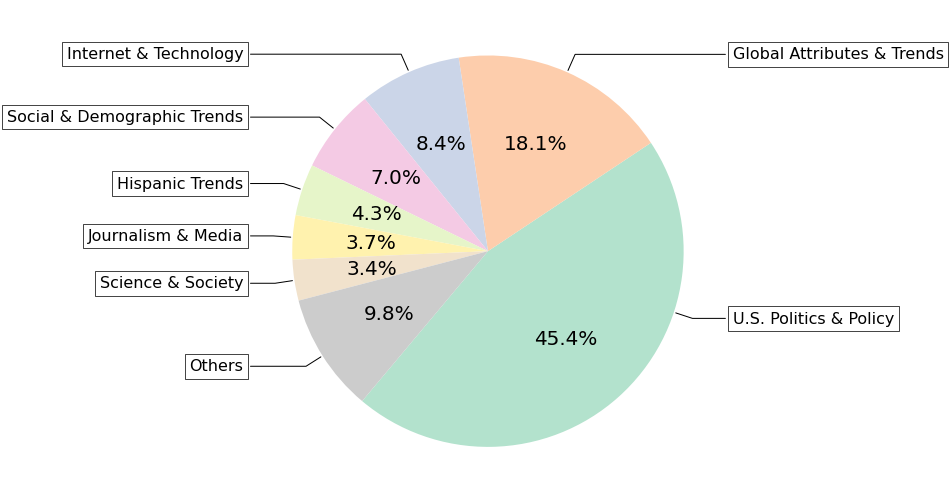}
\caption{\small Pew}
\end{subfigure}
\begin{subfigure}[c]{.49\textwidth}
\centering
    \includegraphics[width=\textwidth]{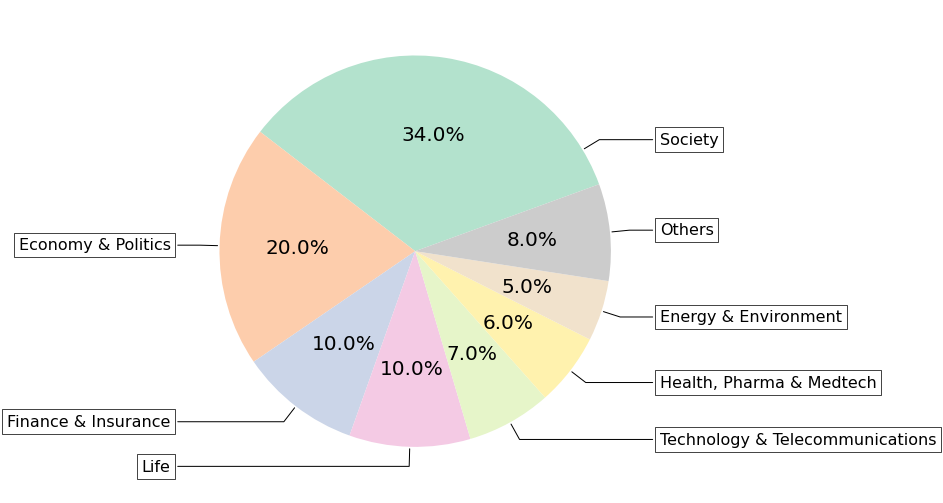}
\caption{\small OECD}
\end{subfigure}
\begin{subfigure}[d]{.49\textwidth}
\centering
    \includegraphics[width=\textwidth]{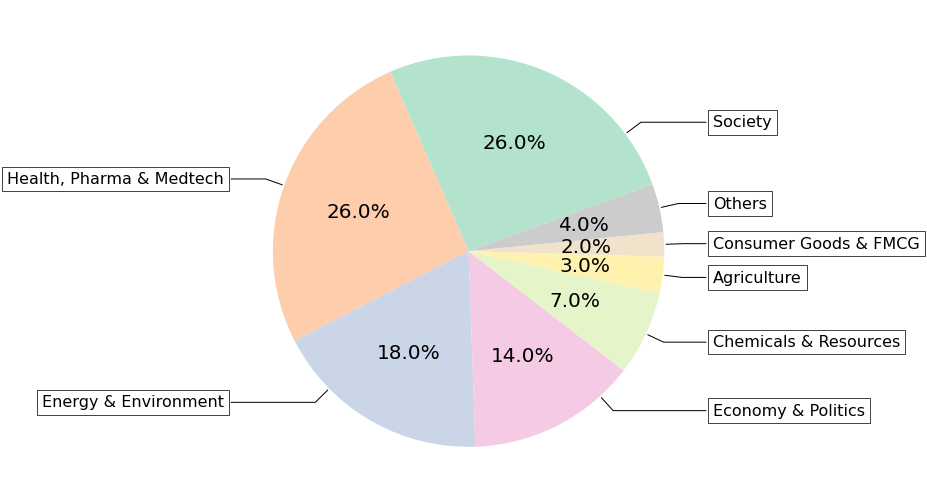}
\caption{\small OWID}
\end{subfigure}
\vspace{-0.5em}
\caption{Distribution of topics in the datasets.}
\label{topics}
\end{figure*}

Table \ref{dataset-token-stats} shows some linguistic statistics about our benchmark. Also, Figure~\ref{topics} shows the distribution of topics in our dataset for each of the four sources. Politics is a common topic among all sources but particularly in the Pew dataset where   nearly half of charts  are about U.S. Politics \& Policy (45.4 \%). The most frequent topic from OECD and OWID is Society (34.0 \% and 26.0 \% respectively).
 
Furthermore, we analyzed how people make visual references to charts in their questions.  Table \ref{visual-ref-stats} shows the usage of visual references made in the randomly selected 300 QA pairs.

\begin{table}[h]
\centering
\small
\scalebox{0.90}{\begin{tabular}{l|c|c}
\toprule
Type  & \textbf{ChartQA-H} & \textbf{ChartQA-M}   \\
\midrule
Avg. Character per question & 60.53   & 67.82 \\
Avg. Character per answer & 5.31   & 5.0 \\
Avg. Token per question & 12.32  & 13.18    \\
Avg. Token per answer & 1.31  & 1.08    \\
Unique tokens in questions & 6,150  & 12,379     \\
Unique tokens in answers & 4,319  & 11,979    \\
Numeric answers & 6,583  & 19,622      \\
Non-numeric answers & 3,025  & 3,489     \\
\bottomrule
\end{tabular}
}
\vspace{-2mm}
\caption{
    \small ChartQA benchmark statistics.
}
\vspace{-2mm}
\label{dataset-token-stats}
\end{table}

 \begin{table}[h!]
\centering
\small
\begin{tabular}{lll}
\toprule
Type  & Examples & Percentage   \\
\hline
Color & green line, red bar & 44.70\% \\
Length & tallest bar & 40.15\% \\
Size & largest pie slice & 11.36\% \\
Position & rightmost, topmost & 8.33\% \\
Counting marks & how many green bars & 3.03\% \\
Unit of a mark & bar unit & 0.76\% \\
\bottomrule
\end{tabular}
\caption{
    Usage of visual references in visual questions
}
\label{visual-ref-stats}
\end{table}

\begin{figure*}[t!]
\centering
    \includegraphics[width=0.95\textwidth,keepaspectratio]{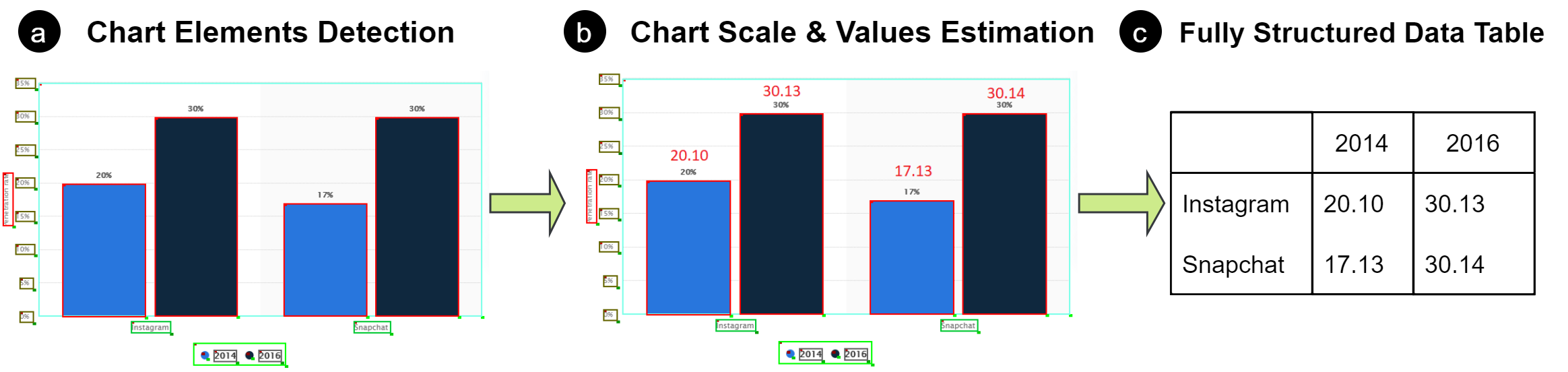}
    \vspace{-3mm}

\caption{\small Data Extraction Process}
\label{fig:data_extraction_example}
\end{figure*}

\begin{figure*}[t!]
\begin{subfigure}[b]{.49\textwidth}
\centering
    \includegraphics[width=1.05\textwidth,keepaspectratio]{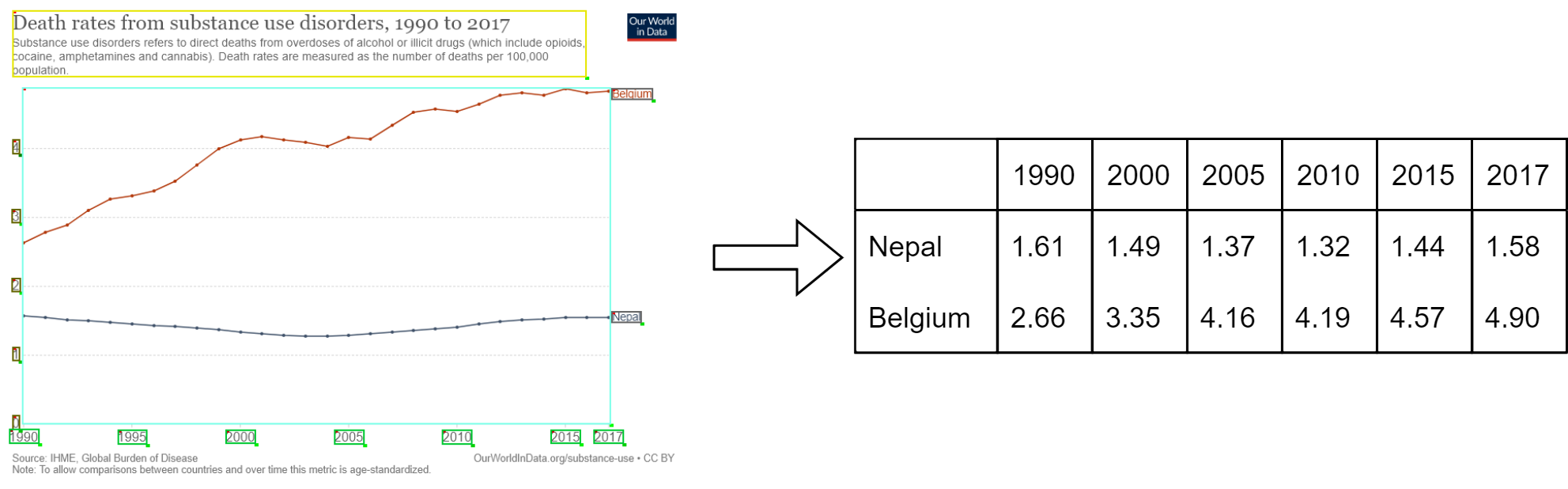}
\caption{\small OWID Line Chart}
\label{fig:data_extraction_owid}
\end{subfigure}
\begin{subfigure}[b]{.49\textwidth}
\centering
    \includegraphics[width=0.80\textwidth,keepaspectratio]{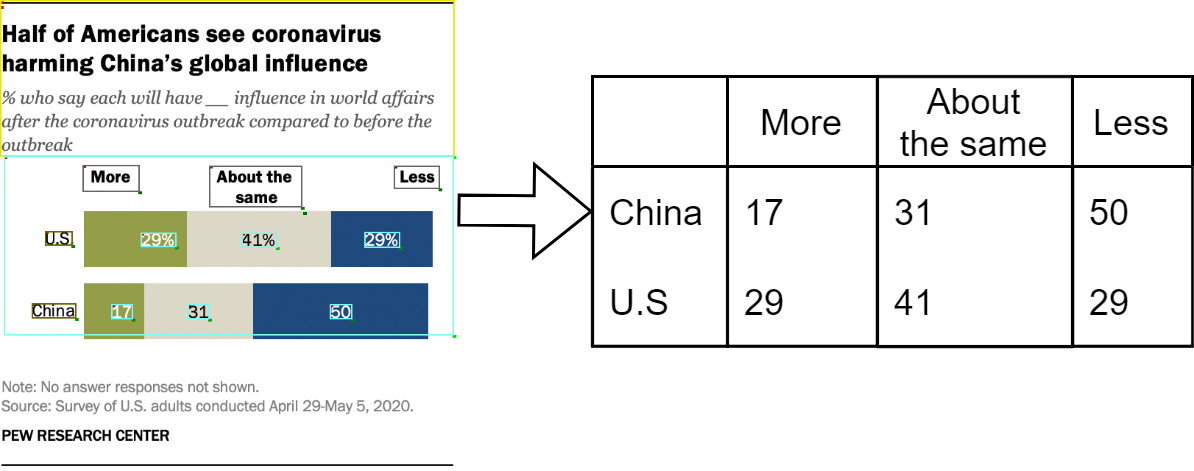}
\caption{\small Pew Bar Chart}
\label{fig:data_extraction_pew}
\end{subfigure}
\vspace{-1em}
\caption{Data extraction examples from OWID and Pew. }
\label{fig:data_extraction_owid_pew}
\end{figure*}

\begin{figure*}[t!]
\begin{subfigure}[b]{.49\textwidth}
\centering
    \includegraphics[width=0.85\textwidth,keepaspectratio]{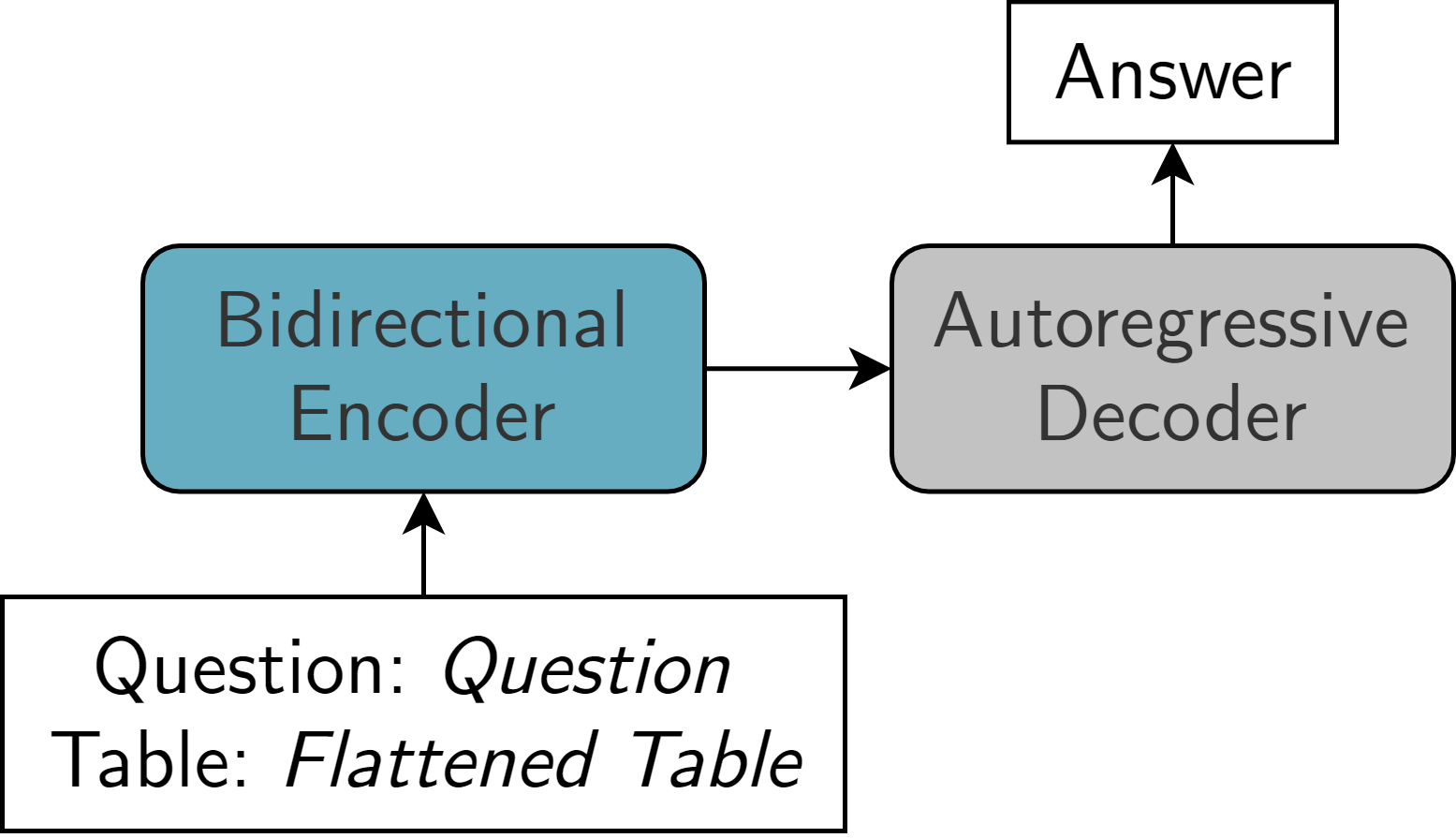}
\caption{\small T5 fine-tuning}
\label{fig:t5}
\end{subfigure}
\begin{subfigure}[b]{.49\textwidth}
\centering
    \includegraphics[width=0.94\textwidth,keepaspectratio]{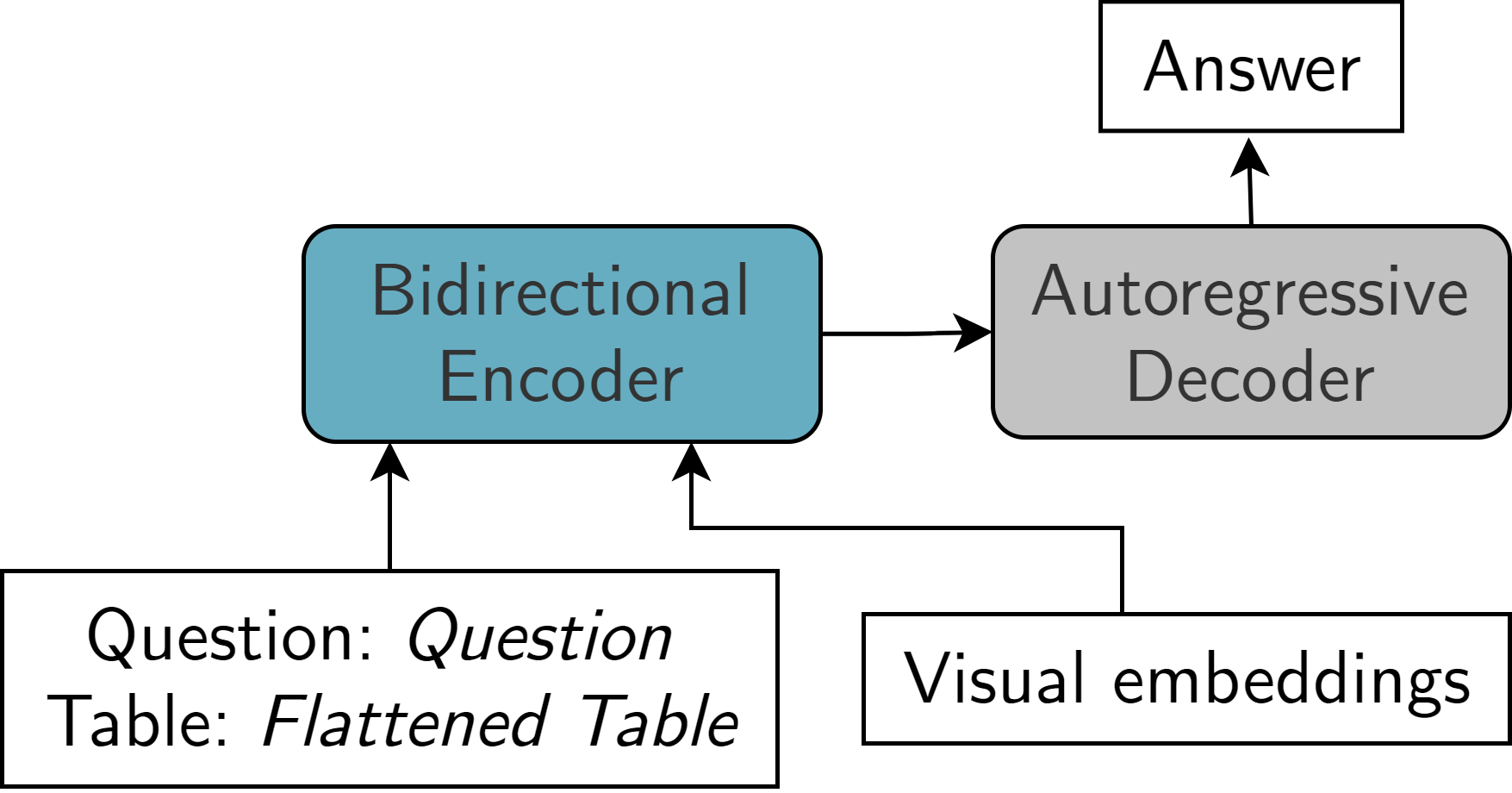}
\caption{\small VL-T5 fine-tuning}
\label{fig:vlt5}
\end{subfigure}
\vspace{-1em}
\caption{Different neural models for ChartQA. Data tables are first flattened and fed into the model along with the question (and visual features in VL-T5). }
\label{fig:t5-vl-t5-models}
\end{figure*}

\subsection{Automatic Chart Data Extraction}
\label{app:data_extraction}
\textbf{Model:} We extend ChartOCR \cite{ChartOCR} which relies on both deep-learning models and rule-based techniques to parse the chart image into the underlying data table. As described in Section (\cref{sec:problem-extraction}), the chart image is parsed in three main stages. In the first stage, key-point detection networks, adapted from \cite{Law2019CornerNetDO}, locates the chart visual marks (\textit{e.g.} bars, plot area, line points). Ideally, the network locates the top-left point and bottom-right points for the rectangular objects (\textit{e.g.} bar, plot area). In line charts, the detection network locates the coordinates of the points connecting the line segments. In pie charts, the network locates the intersection points between the pie segments along the pie perimeter. We extend their detection networks to also locate the chart textual elements (\textit{e.g. x-axis-label, legend-label} ) as shown in Figure \ref{fig:data_extraction_example}a and utilize the CRAFT model \cite{craftrecognition} to read their underlying texts. In the second stage, the chart scale is estimated using the \textit{y-axis-labels} value for line and bar charts, Figure \ref{fig:data_extraction_example}b. For pie charts, the value of each segment is estimated by calculating the angle between its borderlines. Finally, the model aggregates the extracted data values (using color and proximity heuristics) to output the final \textit{raw data values}. We extend their approach to extract the \textit{fully-structured} data table with the textual labels (\textit{e.g.} column headers). As shown in Figure \ref{fig:data_extraction_example}, we associate the estimated bars data values (\textit{e.g.}, `17.13', `40.14') with their closest \textit{x-axis-label} ('Snapchat'). Moreover, if the chart has more than one data series (dark bars or blue bars values), each data series is matched with its \textit{legend-label} (\textit{e.g.}, `2016', `2014') based on the color of the \textit{legend mark} and data-encoding marks (\eg\ bars). 
If we cannot match data values with legends by colors (\eg\ when all legend marks have the same color or there are no legend marks), we use other criteria that associate data-encoding marks with legend marks (\eg\ proximity, alignment). For example, in Figure \ref{fig:data_extraction_pew}, 'More' is matched with '17' and '29' since they are vertically aligned. Similarly, for line charts if there is no explicit legend mark for a line series 
we associate the legend labels with the points of their closest lines as shown in Figure \ref{fig:data_extraction_owid}.

\textbf{Evaluation Metric:} Our evaluation metric is adapted from ChartOCR \cite{ChartOCR}. The distance between any two data values is estimated as follows:
\[
D(gt, pr) = min(1, ||\frac{gt - pr}{gt}||)
\]
where $gt$ is the ground truth value and $pr$ is the predicted value. For each chart, the cost matrix $C$, where $C_{n,m} = D(gt_n, pr_m)$ is computed and the total minimum cost is calculated by solving the following linear sum assignment problem
\[
Cost = \sum_{i=1}^{K}{\sum_{j=1}^{K}{C_{i, j} X_{i,j}}}
\]
Where $K = max(N, M)$ and $X$ is a binary assignment matrix. The final overall score is then estimated as follows:
\[
Overall \ Score = \frac{1}{L}\sum_{i=1}^{L}{1 - \frac{cost}{K_i}}
\]
where $L$ is the total number of charts. Our evaluation results are shown in Table \ref{tab:data_extraction_accuracies}. We have noticed that the accuracy is specifically lower on line and dot line charts in FigureQA and PlotQA. In DVQA, the extracted tables from logarithmic-scale charts were quite noisy since ChartOCR does not support them. Moreover, PlotQA has many charts with very large values (usually written in E notation). Hence, errors in such figures have higher impact on the overall accuracy. Overall, the accuracy on PlotQA and ChartQA are generally lower since they have more complex charts (PlotQA has numerous charts with very large values (\eg\ $1\mathrm{e}^{6}$) and ChartQA has real-world challenging charts). A major limitation of evaluation metrics for the chart data extraction is that they do not take the extracted textual tokens into consideration (which are much more noisy in real-world figures). Hence, better metrics are still needed in the future.  
\begin{table}[h!]
\centering
\scalebox{0.98}{\begin{tabular}{cc}
\toprule
Dataset & \makecell{Accuracy} \\
\hline
FigureQA  & 95.05\%  \\
DVQA & 89.98\% \\
PlotQA & 80.88\% \\
ChartQA & 83.85\% \\
\bottomrule
\end{tabular}}
\caption{
    Accuracies of our data extraction algorithm on the test sets of DVQA, PlotQA, and ChartQA. Since the gold data table is not available in FigureQA, we report the results on the Validation2 set.
}
\label{tab:data_extraction_accuracies}
\end{table}

\subsection{Visual Features Extraction in VL-T5} 
\label{app:maskrcnn}
\paragraph{Object Detection (Mask R-CNN)}
We train the model to detect the following 15 objects: \textit{'Legend'}, \textit{'yAxisTitle'}, \textit{'ChartTitle'}, \textit{'xAxisTitle'}, \textit{'LegendPreview'}, \textit{'PlotArea'}, \textit{'yAxisLabel'}, \textit{'xAxisLabel'}, \textit{'LegendLabel'}, \textit{'PieLabel'}, \textit{'bar'}, \textit{'pie'}, \textit{'pieSlice'}, \textit{'line'}, and \textit{'dotLine'}. For the bounding boxes annotations, we use the available bboxes. For the masks, we generate them easily using the bounding boxes for all the rectangular objects. For \textit{'pieSlice'}  and \textit{'pie'}, we follow a similar approach to \cite{stlcqa} where we generate the masks by projecting the radius along the pie perimeter from the starting to the ending points of each slice. We use the detectron2 library \cite{detectron2} and initialize the model with pre-trained wights on the COCO dataset \cite{coco}. We fine-tune the model with a batch size of 8 and an initial learning rate 0f 0.00025 for 50K iterations.

\subsection{ChartQA Baseline Models}
\label{app:baselines}

T5 and VL-T5 fine-tuning process setup is shown in Figure \ref{fig:t5-vl-t5-models}. 
Our experiments were carried out on one 4-V100 GPU and one 4-A100 GPU machines. Fine-tuning VL-T5 on the PlotQA dataset was the longest experiment which took around 64-70 hours on 4 V100 GPUs. 

\paragraph{TaPas} We follow the same settings as \cite{tapas} on the WikiTQ dataset \cite{sempre} and fine-tune the TaPas-base-wtq for 40K iterations with a batch size 24 on DVQA, PlotQA, and our new dataset. For FigureQA, we follow similar settings to \cite{tabfact} and fine-tune the model with classification objective for 4 epochs with a batch size of 48 and initial learning rate of 0.00001.

\paragraph{VisionTaPas} We fine-tune the model (TaPas-Base 12 layers, ViT-Base 12 layers, and 4 Cross-Modality Layers) for 4 epochs on FigureQA and DVQA, one epoch on PlotQA, and 30 epochs on the new dataset. We use an initial learning rate of 0.00001 and a batch size of 64.

\paragraph{T5} We fine-tune T5-Base (220M, 12 layers) using the huggingface library \cite{huggingface} for 4 epochs on FigureQA, DVQA, and PlotQA datasets and for 30 epochs on our new dataset. We use a batch size of 40 and an initial learning rate of 0.0001. Inference is done with beam search of size 4. 

\paragraph{VL-T5} Similar to T5, we fine-tune VL-T5-Base (220M 12 layers) for 20 epochs on FigureQA and DVQA, 10 epochs on PlotQA, and 30 epochs on our dataset. We use a batch size of 96 and an initial learning rate of 0.0001. Inference is done with beam search of size 5.

\paragraph{PlotQA} We fine-tune the SEMPRE model \cite{sempre} pre-trained on the PlotQA \cite{plotqa} checkpoint for 20 epochs on the new dataset with a batch size of 1 and L1 regularization coefficient of 0.00003.

\paragraph{PReFIL} We follow similar settings 
to \citet{prefil} and train the model for 100 epochs with batch size of 128 and a learning rate of 0.001.

\subsection{Additional Results from Evaluation}
\label{app:results}

Table \ref{tab:chartype} presents the results of two top-performing models in our benchmark by chart types. To analyze question types, we randomly sampled 200 QA pairs from our ChartQA-H and classified them into four main categories.  Table \ref{tab:questiontype} shows the results by question types on this set of 200 QA pairs. 

\begin{table}[h!]
\centering
\scalebox{0.80}{\begin{tabular}{lllll}
\toprule
Model & \makecell{Bar} & \makecell{Line} & \makecell{Pie} & \makecell{Overall} \\
\hline
VisionTaPas  & 49.80\% & 38.20\% & 24.41\% & 45.52\% \\
VL-T5 & 45.82\% & 35.40\% & 25.00\% & 41.56\% \\
\bottomrule
\end{tabular}}
\caption{
    Results for VisionTaPas and VL-T5 on the ChartQA test set by chart type.
}
\label{tab:chartype}
\end{table}

\begin{table}[h!]
\centering
\scalebox{0.6}{\begin{tabular}{cccccc}
\toprule
Model & \makecell{Data \\ Retrieval} & \makecell{Visual \\ Compositional} & \makecell{Compositional} & \makecell{Visual}  & \makecell{Overall} \\
\hline
VisionTaPas  & 60.00\% & 29.78\% & 34.88\% & 16.21\% & 34.00\% \\
VL-T5 & 50.00\% & 19.14\% & 24.41\% & 21.62\% & 26.50\% \\
\bottomrule
\end{tabular}}
\caption{
    Accuracies of VisionTaPas and VL-T5 on the ChartQA-H test set by question type on 200 random samples.
}
\label{tab:questiontype}
\end{table}

\subsection{Sample Questions and Outputs}
\label{app:examples}

\begin{figure*}[t!]
\centering
    \includegraphics[width=0.98\textwidth,keepaspectratio]{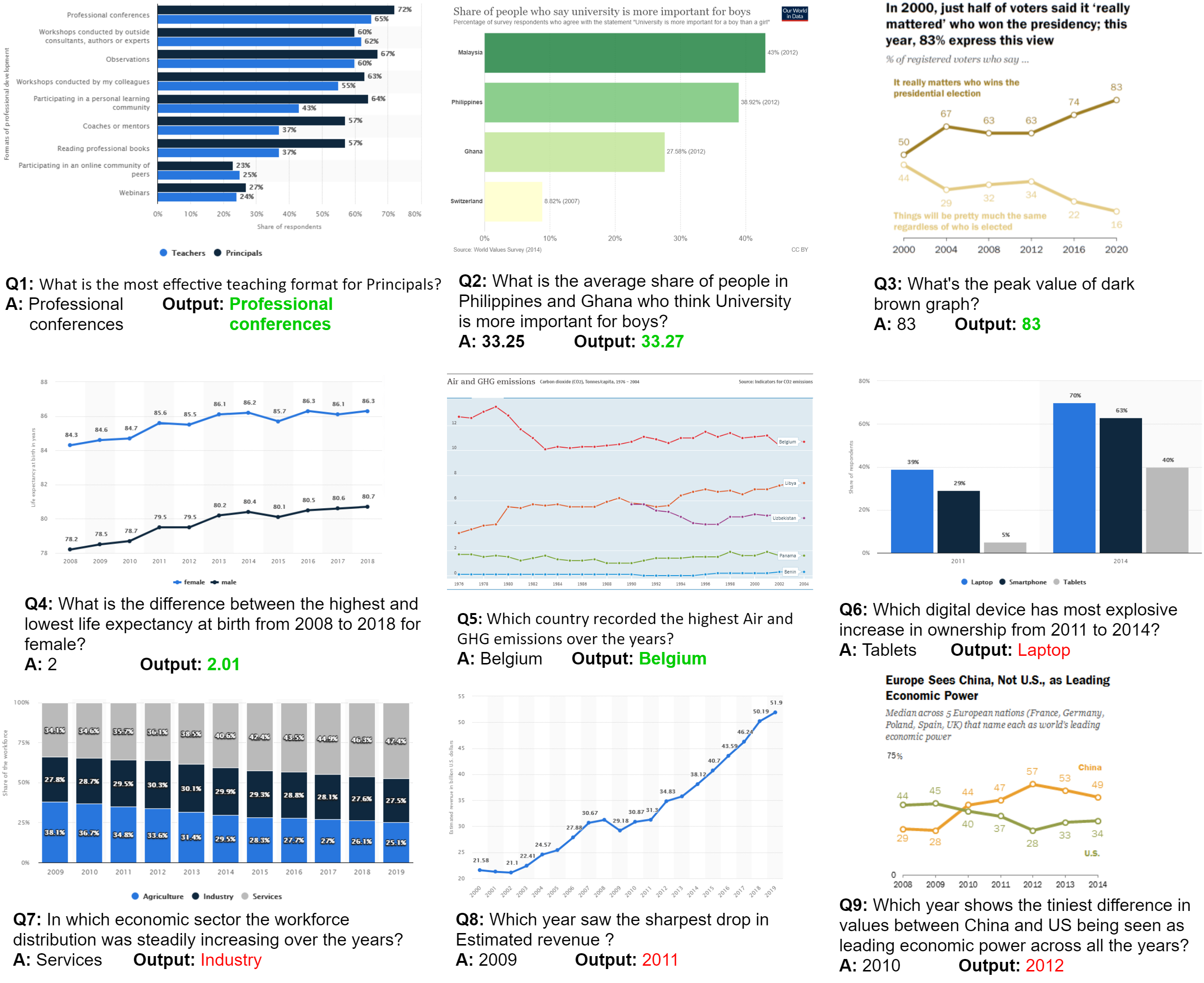}
\caption{\small Sample outputs of our model VisionTaPas on our new ChartQA test set. Answers in green are correct and answers in red are incorrect. }
\label{fig:predictions_examples}
\end{figure*}

Sample machine-generated questions with the human-written summaries are shown in Table \ref{tab:augtmented_examples}.
Sample predictions from our model, VisionTaPas on ChartQA test set are shown in Figure \ref{fig:predictions_examples}.

\begin{table*}[t] 
\linespread{0.5}\selectfont\centering 
\renewcommand{\arraystretch}{1.0}
\scalebox{0.98}{\begin{tabular}{p{1.3cm} | p{9.4cm} | p{3.2cm}|p{0.7cm}}     
\toprule
        {\tiny \textbf{Question Type}}
        &
        {\tiny \textbf{Human-written Summary}}
        &
        {\tiny \textbf{Generated Question}}
        &
        {\tiny \textbf{Answer}}
        \\
\toprule
        {\tiny Compositional}
        &
        {\tiny Cancer was the leading cause of death among state prisoners in the United States, which killed 1,137 state prisoners in 2018. Heart disease was the second leading cause of death in that year, accounting for 1,052 deaths.}
        &
        {\tiny What was the second leading cause of death among state prisoners in 2018?}
        &
        {\tiny Heart disease}
        \\
\midrule 
        {\tiny Compositional}
        &
        {\tiny This statistic shows the number of tourist arrivals at accommodation establishments in Latvia from 2006 to 2019. Since 2009 there has been an increasing trend in arrivals.}
        &
        {\tiny Since what year has there been an increasing trend in arrivals?}
        &
        {\tiny 2009}
        \\
\midrule  
        {\tiny Data Retrieval}
        &
        {\tiny  The statistic shows the youth unemployment rate in the Gambia from 1999 to 2019. According to the source, the data are ILO estimates. In 2019, the estimated youth unemployment rate in the Gambia was at 12.44 percent.}
        &
        {\tiny What was the youth unemployment rate in the Gambia in 2019?}
        &
        {\tiny 12.44 percent}
        \\
\midrule  
        {\tiny Data Retrieval}
        &
        {\tiny  This statistic shows the total population of Portugal from 2016 to 2020, with projections up until 2026. In 2020, the total population of Portugal was at approximately 10.29 million inhabitants.}
        &
        {\tiny In what year did Portugal's population reach 10.29 million?}
        &
        {\tiny 2020}
        \\
 
\bottomrule  
    \end{tabular}
    }
    \vspace{-0.5em}
    \caption{\small Sample question answer pairs generated from human-written summaries in Statista.}
    \label{tab:augtmented_examples}
    \vspace{-4mm}
\end{table*}

\end{document}